\documentclass[letterpaper, 10 pt, conference]{ieeeconf}
\usepackage{cite}
\usepackage{amsmath,amssymb,amsfonts}
\usepackage{graphicx}
\usepackage{hyperref} 
\usepackage{textcomp}
\usepackage{tabularx}
\usepackage{amsmath}
\usepackage{balance}
\usepackage{multirow}
\usepackage{dblfloatfix}
\usepackage{siunitx}
\usepackage{booktabs}
\usepackage{longtable}
\usepackage{cite}
\usepackage{color,colortbl}
\usepackage{algpseudocode}
\usepackage{algorithm}
\definecolor{Gray2}{gray}{0.9}
\definecolor{Gray}{gray}{0.7}

\usepackage{xcolor}
\usepackage{subcaption}

\IEEEoverridecommandlockouts % This command is only needed if 
\begin{document}

% \title{\LARGE \bf AgilePilot: Intelligent Drone Motion Planning with Deep Reinforcement Learning in Dynamic Environments Leveraging Real-time Object Detection\\
% }
\title{\LARGE \bf Adaptive SINDy: Residual Force System Identification Based UAV Disturbance Rejection\\
}

% \author{
% \IEEEauthorblockN{
% Fawad Mehboob, Amir Atef Habel, Roohan Ahmed Khan, Mikhail Derevianchenko \\ Clement Fortin, and Dzmitry Tsetserukou. 
% }
% % \IEEEauthorblockA{
% \thanks{The authors are with the Intelligent Space Robotics Laboratory, Center for Digital Engineering, Skolkovo Institute of Science and Technology, Moscow, Russia. 
% \tt \{Fawad.Mehboob, Amir.Habel, Roohan.Khan, Mikhail.Derevianchenko, c.fortin, d.tsetserukou\}@skoltech.ru}
% \thanks{Research reported in this publication was financially supported by the RSF-DST grant No. 24-41-02039.}
% }
\author{
Fawad Mehboob, Amir Atef Habel, Roohan Ahmed Khan,
Mikhail Derevianchenko, \\
Clement Fortin, and Dzmitry Tsetserukou.
\thanks{$^{1}$Intelligent Space Robotics Laboratory, Center for Digital Engineering, Skolkovo Institute of Science and Technology, Moscow, Russia.
{\tt\{Fawad.Mehboob, Amir.Habel, Roohan.Khan, Mikhail.Derevianchenko, c.fortin, d.tsetserukou\}@skoltech.ru}}%
\thanks{Research reported in this publication was financially supported by the RSF-DST grant No. 24-41-02039.}%
}

\maketitle

\begin{abstract}
The stability and control of Unmanned Aerial Vehicles (UAVs) in a turbulent environment is a matter of great concern. Devising a robust control algorithm to reject disturbances is challenging due to the highly nonlinear nature of wind dynamics, and modeling the dynamics using analytical techniques is not straightforward. While traditional techniques using disturbance observers and classical adaptive control have shown some progress, they are mostly limited to relatively non-complex environments. On the other hand, learning based approaches are increasingly being used for modeling of residual forces and disturbance rejection; however, their generalization and interpretability is a factor of concern. To this end, we propose a novel integration of data-driven system identification using Sparse Identification of Non-Linear Dynamics (SINDy) with a Recursive Least Square (RLS) adaptive control to adapt and reject wind disturbances in a turbulent environment. We tested and validated our approach on Gazebo harmonic environment and on real flights with wind speeds of up to 2 m/s from four directions, creating a highly dynamic and turbulent environment. Adaptive SINDy outperformed the baseline PID and INDI controllers on several trajectory tracking error metrics without crashing. A root mean square error (RMSE) of up to 12.2 cm and 17.6 cm, and a mean absolute error (MAE) of 13.7 cm and 10.5 cm were achieved on circular and lemniscate trajectories, respectively. The validation was performed on a very lightweight Crazyflie drone under a highly dynamic environment for complex trajectory tracking.       

\end{abstract}
{Keywords: UAVs System, Windy Environments, System Identification, Adaptive Control, Disturbance Rejection, Robotics.}

% \vspace{1em}
% \textbf{Video:} 
% \href{https://youtu.be/k8Nf_vPZf7U}{https://youtu.be/k8Nf\textunderscore vPZf7U}

\section{Introduction}
Unmanned Aerial Vehicles (UAVs) have been widely adopted for a variety of applications, including search and rescue missions, delivery of logistics, security and surveillance, agriculture, and much more. UAV researchers are constantly trying to enhance the capacity of these systems to improve flight endurance, speed, payload capacity, and most importantly, their reliability and control stability in harsh and uncertain environmental conditions. One of the biggest challenges for a UAV is flight stability under strong winds. Due to the uncertain nature and high non-linearity of wind dynamics involved, it is extremely difficult to model the uncertainties and devise a control algorithm for safe and reliable flights. Although efforts have been made to model wind as a wind shear model or turbulence models like Dryden and von Karman\cite{shen2023review}, from a control design perspective it is desirable to model a wind invariant representation of the dynamics utilizing the motion of the quadrotor \cite{gonzalez2017measuring}. This leads to modeling of unknown residual forces that apply on the drone in addition to the known UAV dynamics and feedback control input to the system. 

\begin{figure}[t]
  \centering
  \includegraphics[width=\columnwidth]{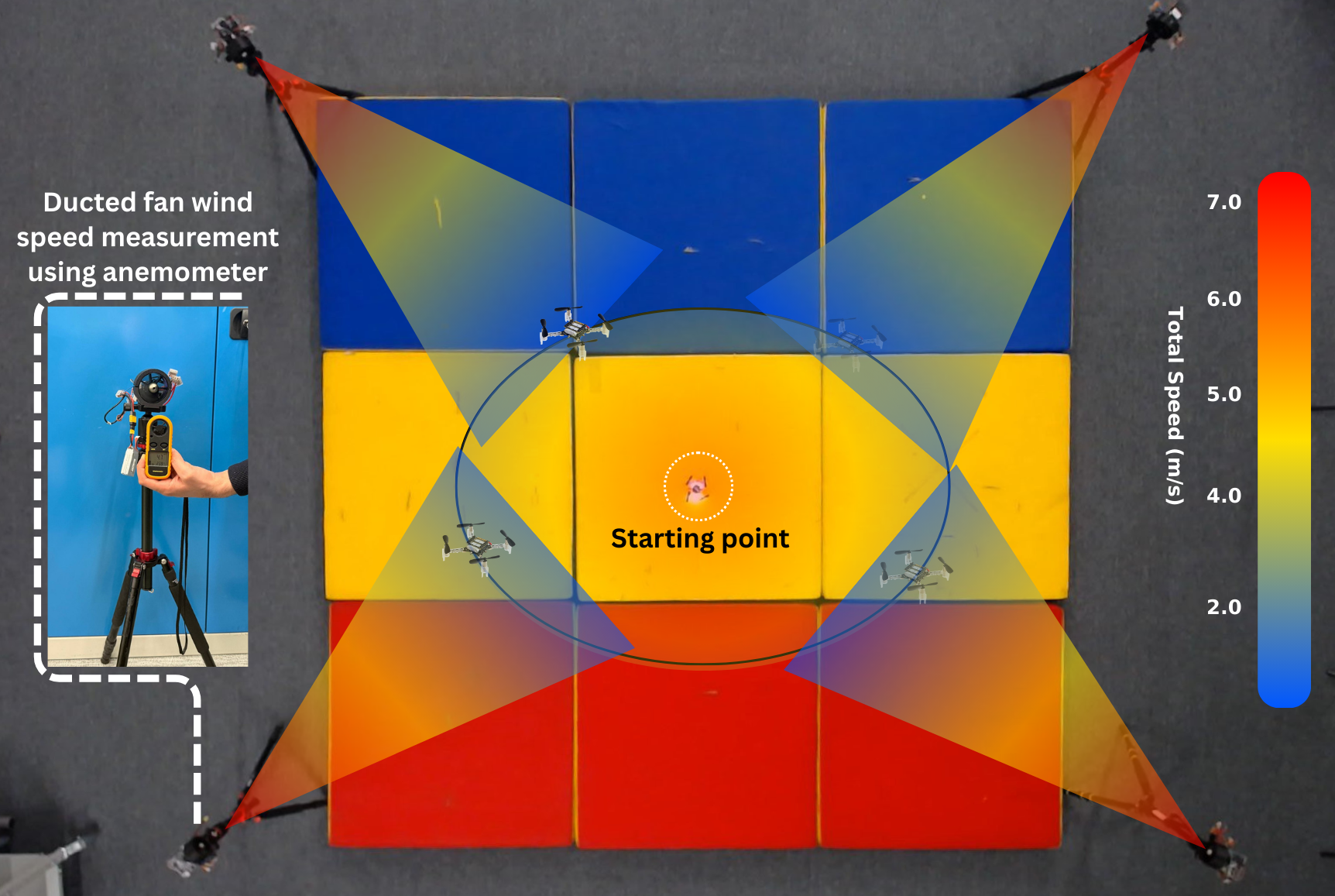}
  \caption{Crazyflie Under Windy Environment, 4 ducted fans blowing air at up to 2 m/s on the drone from four directions.}
  \label{fig:title}
  \vspace{-0.8cm}
\end{figure}

To address the issue of disturbance rejection, researchers have proposed a variety of techniques, including Adaptive Control, Backstepping, Feedback Linearization, Online Self-Tuning PID Control, and more recently, learning based approaches using Neural Networks to model and control the uncertainties. An advanced version of Model Reference Adaptive Control (MRAC) termed as L1 adaptive control was proposed in \cite{fernandez20171}, the approach outperformed the baseline PID and LQR methods. Similarly, a cascaded Incremental Non-Linear Dynamic Inversion (INDI) based controller was proposed in \cite{smeur2023cascaded}, where wind induced dynamic accelerations are incrementally inverted to achieve strong disturbance rejections of wind speeds up to 10 m/s. INDI is also seen in combination with Differential flatness to track aggressive reference trajectories \cite{tal2020accurate}, where the architecture combined the flatness based feedforward acceleration with INDI based trajectory following, achieving accurate tracking under disturbances. Additionally, various disturbance observer approaches are also found in the literature. In \cite{jeong2024control} a variable structure of a disturbance observer is used to account for changing situations, such as addition of a slung load. A disturbance observer based approach for wind disturbance is demonstrated in \cite{wi2025active}, where a generalized active disturbance rejection control was proposed using data-driven model identification. Compared to baseline PID and linear version of disturbance rejection, it showed considerable improvement in performance. Similarly, a non-linear wind disturbance observer is implemented in \cite{9827490} that uses gust frequency to accurately model gust estimation. While all the mentioned techniques have shown significant improvements in drone control under disturbances, challenges still remain in implementation, tuning, and generalizability to name a few.       

The increasing popularity of data-driven modeling and the use of learning from experience approaches have shifted the focus of research from purely classical control techniques to data-driven or hybrid techniques. A framework for the identification of a wind invariant representation of residual forces is presented in \cite{oconnell2022neuralfly}, where pretrained representations of nonlinear disturbance dynamics were learned using a Domain Adversarial Invariant Meta Learning (DAIML) Neural Network, and integrated with adaptive control for disturbance rejection.
% This study leveraged the observation that aerodynamics under different wind conditions can be represented through a common latent representation, while the wind-specific components lie in a low-dimensional subspace. 
% The DAIML enabled effective modeling of aerodynamic variations and combine them with an adaptive control scheme to tune the weights of the learned representation in real time.
Another study used adaptive control with a low-level Reinforcement Learning based control policy for high frequency input commands \cite{zhang2025learning}. A domain invariant adaptive control and RL policy architecture allows implementation of the strategy to different quadrotor models regardless of the system on which it was trained. Similarly, a bio-inspired DRL approach is used in \cite{peter2024tornadodrone} where disturbance rejection is achieved by mimicking the behavior of birds. Additionally, Model Predictive Control (MPC) is also used with learning based approaches for disturbance rejection. In \cite{torrente2021data} aerodynamic residuals are learned using a gaussian process combined with MPC for real time control. 

Although considerable progress has been made in these techniques, each of them has its own strengths, weaknesses, and implementation challenges. Most of the mentioned techniques are system-dependent and lack generalization, while learning based methods require large amounts of experimental data for training and lack interpretability of the resulting models. To address these issues, we propose a novel approach of integrating adaptive control with system identification of disturbance induced residual forces. The non-linear dynamics are identified using Sparse Identification of Non-Linear Dynamics (SINDy) \cite{brunton2016discovering},  which is gaining popularity for discovering hidden dynamics of a system. SINDy requires a relatively small amount of data for training and provides interpretable models for the discovery of dynamics. The main advantage of using SINDy is the flexibility in choosing the basis functions to model unknown dynamics. Additionally, integration of control inputs into the optimization structure of SINDy allows identifying dynamics of unknown systems with exogenous inputs \cite{brunton2016sparse}. Moreover, the sparsity promoting optimization allows for identification of a parsimonious representation of the residual forces. SINDy has been applied to many dynamical systems, but its success highly depends on several design choices as discussed in \cite{pandey2024data}. Here, we use a novel framework using a tailored library of basis functions, data pre-processing, and model learning to address these issues.  
% SINDy has the ability to identify a sparse combination of terms for the dynamical system and is easily scalable to high dimensional states with a relatively small amount of data required. It was shown by \cite{pandey2024data} that a combination of steady-state responses and transient responses are crucial in identifying the correct model using SINDy. 
% Moreover, the design choice for the feature library, choice of optimizers, regularization and thresholding parameters, amount of noise, and the amount of data also greatly impact the performance in system identification. To further delineate the effects of these factors, a benchmarking analysis was presented by \cite{kaptanoglu2023benchmarking} where several SINDy optimizers were compared over 70 chaotic systems swept over a range of possible hyperparameter.
 To the best of our knowledge, the application of SINDy based system identification for UAVs disturbance rejection is largely unexplored, and in this study we aim at demonstrating a system identification framework coupled with adaptive control for online disturbance rejection of quadrotor UAVs under disturbances. Combining the interpretability of SINDy with the online adaptation capability of adaptive control, this study presents a robust framework for reliable and safe UAVs flights under strong wind disturbances.

\section{Methodology}

\subsection{Sparse Identification of Non-Linear Dynamics (SINDy)}

\begin{figure*}[t]
  \centering
  \includegraphics[width=\textwidth]{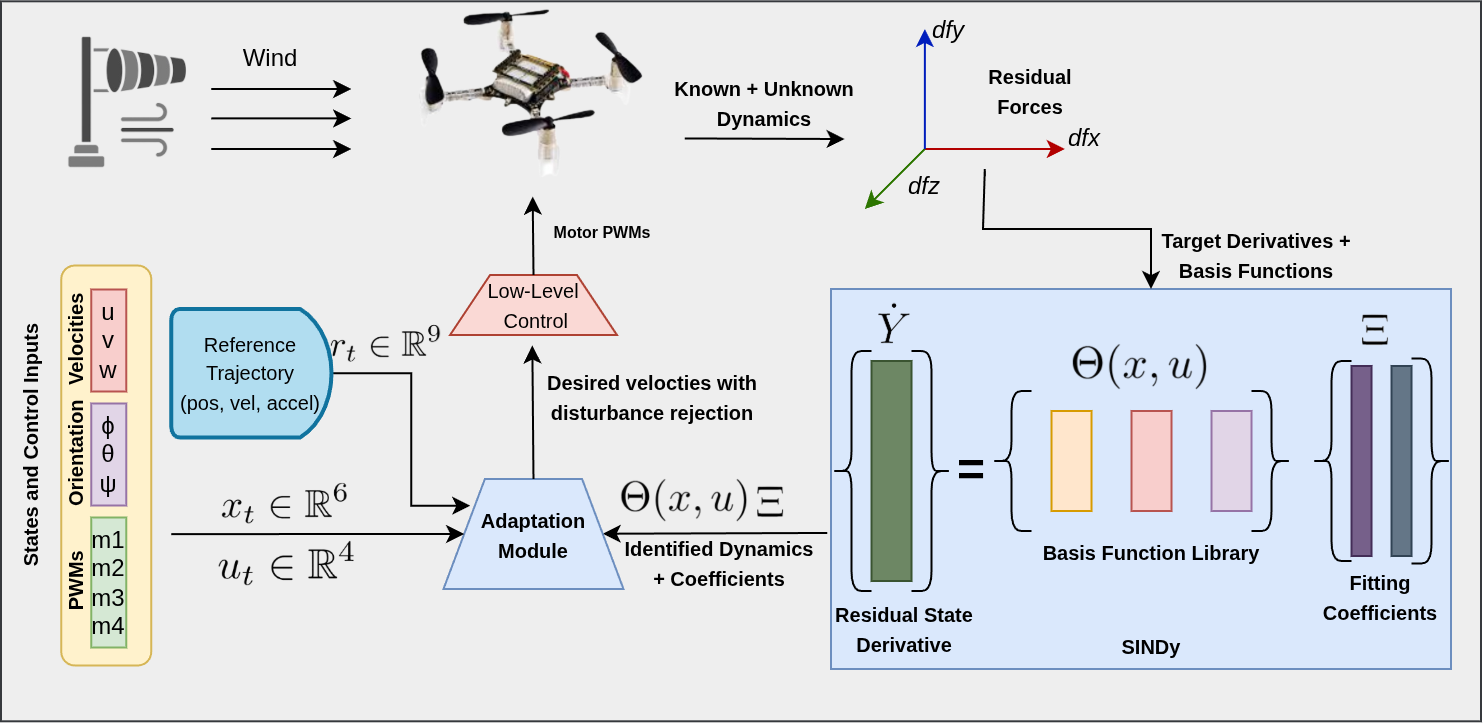}
  \caption{Adaptive SINDy system architecture for system identification based residual force estimation coupled with RLS adaptive control to reject disturbance.}
  \label{fig:architecture}
\end{figure*}

The use of Sparse Identification of Nonlinear Dynamics (SINDy) to identify nonlinear dynamics associated with wind modeling has been demonstrated in several studies \cite{lee5701416sparse,babu2023identification}. However, system identification based adaptation for UAVs disturbance rejection is largely unexplored. This research proposes a novel integration of adaptive control with the online identification of nonlinear dynamics. A high-level overview of the proposed approach is illustrated in Fig. \ref{fig:architecture}.

When a UAV is subject to wind disturbances, its dynamics can be modeled as a combination of known and unknown components:

\begin{equation}
\dot{X} = f(x,u)_{known} +  g(x,u,w)_{unknown},
\label{knwon_unknown}
\end{equation}

Here, $x \in \mathbb{R}^{12}$ and $u \in \mathbb{R}^{4}$ denote the system states and control inputs, respectively, while $w$ represents the unknown wind disturbance. The overall dynamics are decomposed into known dynamics $f(x,u)$ and unknown disturbance-induced dynamics $g(x,u,w)$. $\dot{X} \in \mathbb{R}^{12}$ represents the state derivatives of the 12 drone states including the positions, velocities, angular velocities, and orientation logged during a flight for data collection. A target vector $\dot{Y}$ is defined to represent the difference between the actual dynamics of the system and the known model, as the state derivative of the residual forces:

\begin{equation}
\dot{Y} = \dot{X} - f(x,u)_{known},
\label{Target_Difference}
\end{equation}

The SINDy algorithm identifies a sparse set of coefficients $\Xi$ by weighting a library of candidate functions $\Theta(x,u)$ such that the following relation is satisfied \cite{brunton2016discovering}:

\begin{equation}
\dot{Y} =  \Theta(x,u) \, \Xi.
\label{Sindy}
\end{equation}

The library matrix $\Theta(x,u)$ may consist of polynomial, trigonometric, or exponential functions of the system states and inputs, which are assumed to capture the relevant dynamics. SINDy employs sparse regression to identify the most significant terms in this library. Once the unknown dynamics are identified, an adaptive control algorithm performs online tuning of the coefficients $\Xi$ to drive the system towards the desired states.

% The central hypothesis of this work is that adaptive control performance can be significantly enhanced when model uncertainties are accurately approximated. During nominal, disturbance-free flight conditions, the known dynamics closely approximate the true system behavior, making control relatively straightforward. In such cases, the SINDy-identified dynamics remain inactive, and the adaptive controller reduces the influence of the coefficients in $\Xi$. However, when the system experiences external disturbances, the adaptation mechanism becomes active, and the coefficients $\Xi$ increase accordingly. These updated coefficients are then used to compute the desired system rates, which are tracked by the onboard controllers to compensate for the disturbances in real time.

Although the flexibility to choose the basis functions is advantageous in allowing us to choose the physics informed functions, curating the terms according to the relevant physics in terms of the drone states instead of the wind is challenging. A wind invariant representation of the residual forces solely from drone states is best captured using neural networks as they give a map from inputs to outputs, but they are not interpretable in terms of describing the physics of the system. However, one of the main observations in simulating a drone flight under wind disturbance in the Gazebo simulator was an induced angle of tilt in the opposing direction of the wind. This is due to the inbuilt controller fighting the wind to track position; since the wind is constantly opposing the controller's effort to track the position, the controller induces a tilt towards the opposing wind thus creating roll and pitch angles. This observation is used to create a library of basis functions that includes the pitch angle $\theta$, roll angle $\phi$, and thrust weighted sinusoids of these angles as:
% \begin{equation}
% \small
% \Theta (x,u) = 
% \begin{bmatrix}
% | & | & | & | & | & | & | \\  
% 1 & \theta & \phi & Tsin(\theta) & T cos(\theta) & Tsin(\phi) & T cos(\phi) \\
% | & | & | & | & | & | & | 
% \end{bmatrix}.
% \end{equation}
% % preamble
\begin{equation}
\begingroup
\setlength{\arraycolsep}{1.5pt} % tighter column spacing inside matrix
\renewcommand{\arraystretch}{0.85}
\resizebox{\columnwidth}{!}{$
\Theta(x,u)=
\begin{bmatrix}
\vert & \vert & \vert & \vert & \vert & \vert & \vert \\
1 & \theta & \phi & T\sin(\theta) & T\cos(\theta) & T\sin(\phi) & T\cos(\phi) \\
\vert & \vert & \vert & \vert & \vert & \vert & \vert
\end{bmatrix}.
$}
\endgroup
\end{equation}
The SINDy formulation is well-posed for an open loop system without any state dependent feedback. If the input to the system is dependent on the states such as $u = f(x)$, identifying the true dynamics of the system becomes complex, as it requires delineating the input terms as feedback state inputs treating the system as open loop. Fortunately, we consider the residual inputs as exogenous inputs due to the wind and bypass the closed-loop complexity of the flight controller. Several choices for optimization algorithms are available for the SINDy algorithm. The most commonly used are the Sequentially Thresholded Least Squares (STLSQ), Mixed-Integer Optimized Sparse Regression (MIOSR), and Sparse Relaxed Regularized Regression (SR3) to name a few. We have used the SR3 due to its ability to apply constraints to the optimization problem:

\begin{equation}
\begin{split}
\underset{w,u}{min} \quad \frac{1}{2}|| \dot{Y} - \Theta w||_2^2 +\lambda R(u)+\frac{1}{2\nu} ||w - u||_2^2, \\
s.t. \quad Cw = d,
\end{split}
\end{equation}
where $w$ represents the fitting coefficients, $ u$ is an auxiliary coefficient vector, $\lambda$ is the sparse regularizer, $\nu$ is the relaxation parameter, and $C$ is the set of constraints. SR3 decouples model fitting and sparsification by introducing $u:w$ fits $\dot{Y} \approx \Theta w$, while u is driven sparse by $R(u)$. The relaxation term $\frac{1}{2\nu} ||w - u||_2^2$ links them together, improving numerical stability and enabling constrained sparse identification via $Cw = d$, where $d$ is a vector of numerical values.

% \begin{figure*}[htbp]
%   \centering
%   \includesvg[width=0.9\linewidth]{images/Adaptive_Sindy_architecture}
%   \caption{Adaptive SINDy system Architecture.}
% \end{figure*}

\subsection{Adaptive Control Algorithm}

The controller takes in a reference trajectory of position, velocity, and acceleration as $r(t) \in \mathbb{R}^9$. The commanded acceleration is computed using PD tracking errors in position and velocity:
\begin{equation}
a_{cmd} = a_d - K_p e_p - K_v e_v.
\end{equation}

The actual acceleration $a_{meas}$ representing the residual forces is computed using the finite difference of the measured velocities. This is used in computing a residual force proxy estimation using:

\begin{equation}
f_{dist} = m a_{meas} + mg_{vec} - u_{prev},
\end{equation}
where $m$ is the mass of the drone, $u_{prev}$ is the thrust vector expressed in the world frame, obtained from the estimated thrust magnitude and attitude. This is followed by computing the difference between the computed residual force and an estimated residual force denoted as $\hat{f}_{dist} = \phi A$, where $ A \in \mathbb{R}^{n\times3}$ is the adaptation parameter matrix, and $\phi$ is the set of $n$ active library terms of SINDy. 

The adaptation update of the $A$ matrix is performed using an RLS leaky continuous time update law:
\begin{equation}
\dot{A} = -\lambda A - P \phi^T (\frac{\hat{f}_{dist} - f_{dist}}{R}) + P \phi^T s,
\end{equation}
where P is the covariance matrix, R is measurement error matrix, $\lambda$ is the leakage weight, and $s$ is the sliding error term. The covariance update is then computed as:

\begin{equation}
\dot{P} = -2 \lambda P + Q - \frac{1}{\bar{R}}P (\phi^T)\phi.
\end{equation}

Finally, the desired force is computed using the difference between the commanded force and estimated residual as:

\begin{equation}
F_d = m a_{cmd} + mg_{vec} - \hat{f}_{dist}. 
\end{equation}

This desired force is then used to compute the required thrust and attitude using kinematics. The desired thrust and attitude are commanded to the low level flight controller. Algorithm \ref{alg:adaptive_sindy_short} delineates the adaptive control algorithm.

\begin{algorithm}[t]
\caption{Adaptive SINDy Disturbance Compensation}
\label{alg:adaptive_sindy_short}
\begin{algorithmic}[1]
\Require Reference $(p_d,v_d,a_d)$, gains $K_p,K_v,\Lambda$, mass $m$, gravity $g_{\text{vec}}$
\Require Feature map $\phi(\cdot)$, initial $A$, covariance $P$, parameters $\lambda,Q,R$
\For{each control step}
\State Read $p,v,R_{wb}$ and estimate thrust vector $u_{prev}$
\State $e_p \gets p-p_d,\;\; e_v \gets v-v_d$
\State $a_{cmd} \gets a_d - K_p e_p - K_v e_v$
\State Estimate $a_{meas}$ from velocity difference (optionally low-pass filter)
\State $f_{dist} \gets m a_{meas} + m g_{\text{vec}} - u_{prev}$
\State $\hat f_{dist} \gets \phi\,A$
\State $s \gets e_v + \Lambda e_p$
\State Update $A$ and $P$ using leaky RLS with error $(\hat f_{dist}-f_{dist})$ and term $s$
\State $F_d \gets m a_{cmd} + m g_{\text{vec}} - \hat f_{dist}$
\State Convert $F_d$ to desired thrust and attitude and send to the flight controller
\EndFor
\end{algorithmic}
\end{algorithm}

\begin{table*}[h]
\centering
\caption{Ardupilot SITL tracking error statistics across trajectories (mean $\pm$ std over 10 runs).}
\label{tab:ardupilot_error}
\scriptsize
\setlength{\tabcolsep}{16pt}
\begin{tabular}{|c|c|c|c|c|c|}
\hline
\textbf{Trajectory} & \textbf{Control} & \textbf{RMSE$_{xy}$} & \textbf{MAE$_{xy}$} & \textbf{P95$_{\|xy\|}$} & \textbf{Max$_{\|xy\|}$} \\
 &  & \textbf{[m]} & \textbf{[m]} & \textbf{[m]} & \textbf{[m]} \\
\hline

\multirow{3}{*}{Circle}
 & ADAIML & \textbf{0.251 $\pm$ 0.072} & \textbf{0.195 $\pm$ 0.029} & \textbf{0.514 $\pm$ 0.242} & \textbf{0.851 $\pm$ 0.420} \\
\cline{2-6}
 & \textbf{ASINDy} & 0.307 $\pm$ 0.064 & 0.234 $\pm$ 0.025 & 0.635 $\pm$ 0.232 & 1.170 $\pm$ 0.360 \\
\cline{2-6}
 & PID & 0.503 $\pm$ 0.515 & 0.343 $\pm$ 0.257 & 1.213 $\pm$ 1.552 & 1.658 $\pm$ 1.887 \\
\hline

\multirow{3}{*}{Infinity}
 & ADAIML & 0.373 $\pm$ 0.352 & 0.267 $\pm$ 0.181 & 0.837 $\pm$ 0.989 & 1.395 $\pm$ 1.374 \\
\cline{2-6}
 & \textbf{ASINDy} & \textbf{0.350 $\pm$ 0.083} & \textbf{0.262 $\pm$ 0.054} & \textbf{0.773 $\pm$ 0.200} & \textbf{1.231 $\pm$ 0.340} \\
\cline{2-6}
 & PID & 0.652 $\pm$ 0.571 & 0.434 $\pm$ 0.296 & 1.639 $\pm$ 1.723 & 2.308 $\pm$ 2.059 \\
\hline

\multirow{3}{*}{Spiral}
 & ADAIML & \textbf{0.481 $\pm$ 0.219} & \textbf{0.287 $\pm$ 0.108} & \textbf{0.972 $\pm$ 0.686} & \textbf{2.193 $\pm$ 0.804} \\
\cline{2-6}
 & \textbf{ASINDy} & 0.659 $\pm$ 0.328 & 0.373 $\pm$ 0.126 & 1.622 $\pm$ 1.096 & 2.774 $\pm$ 1.409 \\
\cline{2-6}
 & PID & 1.490 $\pm$ 1.988 & 0.730 $\pm$ 0.821 & 4.099 $\pm$ 5.822 & 6.335 $\pm$ 9.368 \\
\hline

\end{tabular}
\end{table*}

% Amir edited 

\subsection{Implementation Details}
The proposed framework is evaluated on two systems: (i) Gazebo Harmonic simulation of the \texttt{iris\_runway} environment connected to ArduPilot Software-In-The-Loop (SITL) through MAVROS, and (ii) Gazebo Harmonic simulation of the Crazyflie using the CrazySim firmware stack. For both systems, three reference trajectories (circle, lemniscate/infinity, and spiral) are tracked under wind disturbances. During each run, the reference trajectory and the estimated state (position, velocity, attitude, and angular rates) are logged and later used to compute tracking error statistics.

We compare Adaptive SINDy against an adaptive controller based on a pretrained DAIML disturbance estimation model \cite{oconnell2022neuralfly}, and a baseline PID on both the systems. For Crazyflie, an additional INDI controller is also evaluated. Each method is executed for multiple independent runs per trajectory (10 runs for most cases), and the reported statistics are summarized as mean $\pm$ standard deviation in all runs.

\paragraph{Wind disturbance injection (external wrench)}\mbox{}\\
To emulate wind gusts in simulation, an external force (wrench) is applied directly to the vehicle body in the world (ENU) frame using Gazebo's wrench interface.
Wind is applied in bursts using an on/off schedule. Let $T_\text{on}$ and $T_\text{off}$ denote the duration of gust on and off, and define the cycle period $T_c = T_\text{on} + T_\text{off}$ with a binary indicator $\delta(t)\in\{0,1\}$, the gust application is then defined as:
\begin{equation}
\delta(t)=
\begin{cases}
1, & \text{if } (t \bmod T_c) < T_\text{on},\\
0, & \text{otherwise.}
\end{cases}
\end{equation}
When $\delta(t)=1$ the gust is active, and when $\delta(t)=0$ the disturbance smoothly decays to its mean.

\paragraph{Ornstein--Uhlenbeck (OU) gust model}\mbox{}\\
The stochastic gust component $g(t)\in\mathbb{R}^3$ is modeled as an Ornstein--Uhlenbeck process \cite{pham2025application}:
\begin{equation}
dg(t) = \theta(\mu - g(t))\,dt + \sigma\,dW(t),
\label{eq:OU_process}
\end{equation}
where $\mu\in\mathbb{R}^3$ is the mean (drift) vector, $\theta>0$ is the mean reversion rate, $\sigma>0$ controls gust intensity, and $W(t)$ is standard Brownian motion.
Eq. \ref{eq:OU_process} is discretized using an Euler discretization with step $\Delta t$ and a fixed random seed is used for reproducibility of the gust realization for the stochastic $W(t)$ process.

\paragraph{Wind force composition and application}\mbox{}\\
The commanded wind disturbance is generated by combining a slow-varying sinusoidal component with the stochastic OU gust term, resulting in a raw force
\begin{equation}
F_\text{raw}(t)=
\begin{bmatrix}
f_{x,\text{mean}} + f_{x,\text{amp}}\sin(2\pi f t) + g_x(t)\\
f_{y,\text{mean}} + f_{y,\text{amp}}\sin(2\pi f t + \phi_0) + g_y(t)\\
f_{z,\text{mean}} + g_z(t)
\end{bmatrix},
\end{equation}
where $f_{xyz,mean}$ is the mean amplitude of the forces, $f_{xy,amp}$ is the max amplitude of the forces, $f$ is the frequency of the sinosoids, and $\phi_0$ is the phase difference.

To avoid unrealistically large impulses, the force is then limited in magnitude and rate-of-change to obtain the final applied wrench force $F_\text{wind}(t)$. For logging and interpretation, the equivalent wind acceleration is computed as:
\begin{equation}
a_\text{wind}(t)=\frac{1}{m}F_\text{wind}(t).
\end{equation}

\paragraph{Data Collection}\mbox{}\\
The data collection was performed on Gazebo Harmonic simulation in the \texttt{iris\_runway} environment on ArduPilot SITL through mavros. To identify the dynamics it is important to have a rich dataset containing excitation of multiple modes of the system. Since the wind is applied as an external input to the system, we focus on applying a random OU process wind force while tracking trajectories like circle and lemniscate. All trajectories including drone states and control inputs are logged using mavros topics by collecting Rosbags. The reference and achieved trajectories are logged at approximately 40 Hz and a simple pre-processing is performed by smoothing the data using a running mean and severing off the edges to focus mainly on trajectory tracking. SINDy model is trained on the difference computed from the known and unknown dynamics using Eq. \ref{Target_Difference} and \ref{Sindy} and used as a representation of the residual forces in Eq. \ref{alg:adaptive_sindy_short}. 

\paragraph{Experiment protocol and error metrics}\mbox{}\\
For each run, the vehicle is commanded to follow the reference trajectory while the disturbance is applied using $F_\text{wind}(t)$ at a fixed update rate in the simulation, and with ducted fans in real flight. The planar position tracking error is defined as $e_{xy}(t) = p_{xy}(t)-p^{\text{ref}}_{xy}(t)$ and we report Root Mean Square Error (RMSE) and a tail metric based on the 95th percentile of $\|e_{xy}(t)\|_2$ over the runs. The final performance is summarized as a mean $\pm$ standard deviation across multiple runs for each controller and trajectory.
For real flight experiments, a Crazyflie drone is used inside a turbulent wind environment where the drone experiences wind from all four sides. Ducted fans are used to blow air on the drone at speeds of 1-2 m/s (1.2 m/s in the center, increasing to 2 m/s at the edge). The drone is commanded to follow three types of trajectories and its response to gusts from all sides is observed.

\section{Results}

The validation of the Adaptive SINDy disturbance rejection was performed on Gazebo simulation and real experiments on three trajectories i.e., circle, lemniscate, and spiral. The proposed approach was compared with a pre-trained DAIML based adaptive controller, and a baseline PID on two different systems. A simulation based analysis was performed on Gazebo + ArduPilot SITL environment, and a Gazebo + Crazyflie environment was also tested. Additionally, the validation was performed on Crazyflie in real experiments as well. The following sections summarize the results by comparing the error statistics computed on multiple runs of each controller.

\begin{table*}[t]
\centering
\caption{Crazyflie tracking error statistics across trajectories (mean $\pm$ std over 10 runs).}
\label{tab:crazyflie_tracking_xy}
\scriptsize
\setlength{\tabcolsep}{16pt}
\begin{tabular}{|c|c|c|c|c|c|}
\hline
\textbf{Trajectory} & \textbf{Control} & \textbf{RMSE$_{xy}$} & \textbf{MAE$_{xy}$} & \textbf{P95$_{\|xy\|}$} & \textbf{Max$_{\|xy\|}$} \\
 &  & \textbf{[m]} & \textbf{[m]} & \textbf{[m]} & \textbf{[m]} \\
\hline

\multirow{4}{*}{Circle}
 & ADAIML & \textbf{0.216 $\pm$ 0.004} & 0.154 $\pm$ 0.007 & \textbf{0.248 $\pm$ 0.007} & 1.100 $\pm$ 0.000 \\
\cline{2-6}
 & \textbf{ASINDy} & 0.241 $\pm$ 0.001 & \textbf{0.133 $\pm$ 0.004} & 0.577 $\pm$ 0.005 & 1.100 $\pm$ 0.000 \\
\cline{2-6}
 & INDI & 0.260 $\pm$ 0.004 & 0.176 $\pm$ 0.004 & 0.558 $\pm$ 0.011 & \textbf{1.083 $\pm$ 0.034} \\
\cline{2-6}
 & PID & 0.297 $\pm$ 0.031 & 0.222 $\pm$ 0.046 & 0.600 $\pm$ 0.043 & 1.102 $\pm$ 0.006 \\
\hline

\multirow{4}{*}{Infinity}
 & ADAIML & 0.128 $\pm$ 0.009 & 0.116 $\pm$ 0.012 & 0.218 $\pm$ 0.018 & \textbf{0.241 $\pm$ 0.017} \\
\cline{2-6}
 & \textbf{ASINDy} & \textbf{0.105 $\pm$ 0.009} & \textbf{0.092 $\pm$ 0.011} & \textbf{0.161 $\pm$ 0.008} & 0.304 $\pm$ 0.003 \\
\cline{2-6}
 & INDI & 0.140 $\pm$ 0.026 & 0.127 $\pm$ 0.026 & 0.223 $\pm$ 0.032 & 0.297 $\pm$ 0.077 \\
\cline{2-6}
 & PID & 0.189 $\pm$ 0.041 & 0.171 $\pm$ 0.037 & 0.299 $\pm$ 0.068 & 0.352 $\pm$ 0.051 \\
\hline

\multirow{4}{*}{Spiral}
 & ADAIML & 0.119 $\pm$ 0.023 & 0.107 $\pm$ 0.024 & 0.190 $\pm$ 0.026 & 0.211 $\pm$ 0.006 \\
\cline{2-6}
 & \textbf{ASINDy} & \textbf{0.072 $\pm$ 0.003} & \textbf{0.063 $\pm$ 0.003} & \textbf{0.121 $\pm$ 0.004} & \textbf{0.132 $\pm$ 0.005} \\
\cline{2-6}
 & INDI & 0.125 $\pm$ 0.006 & 0.111 $\pm$ 0.007 & 0.208 $\pm$ 0.006 & 0.223 $\pm$ 0.006 \\
\cline{2-6}
 & PID & 0.176 $\pm$ 0.049 & 0.161 $\pm$ 0.047 & 0.272 $\pm$ 0.068 & 0.299 $\pm$ 0.077 \\
\hline

\end{tabular}
\end{table*}

\subsection{ArduPilot SITL Disturbance Rejection}

\begin{figure}[htbp]
  \centering

  \begin{subfigure}[b]{0.42\textwidth}
    \centering
    \includegraphics[width=\linewidth]{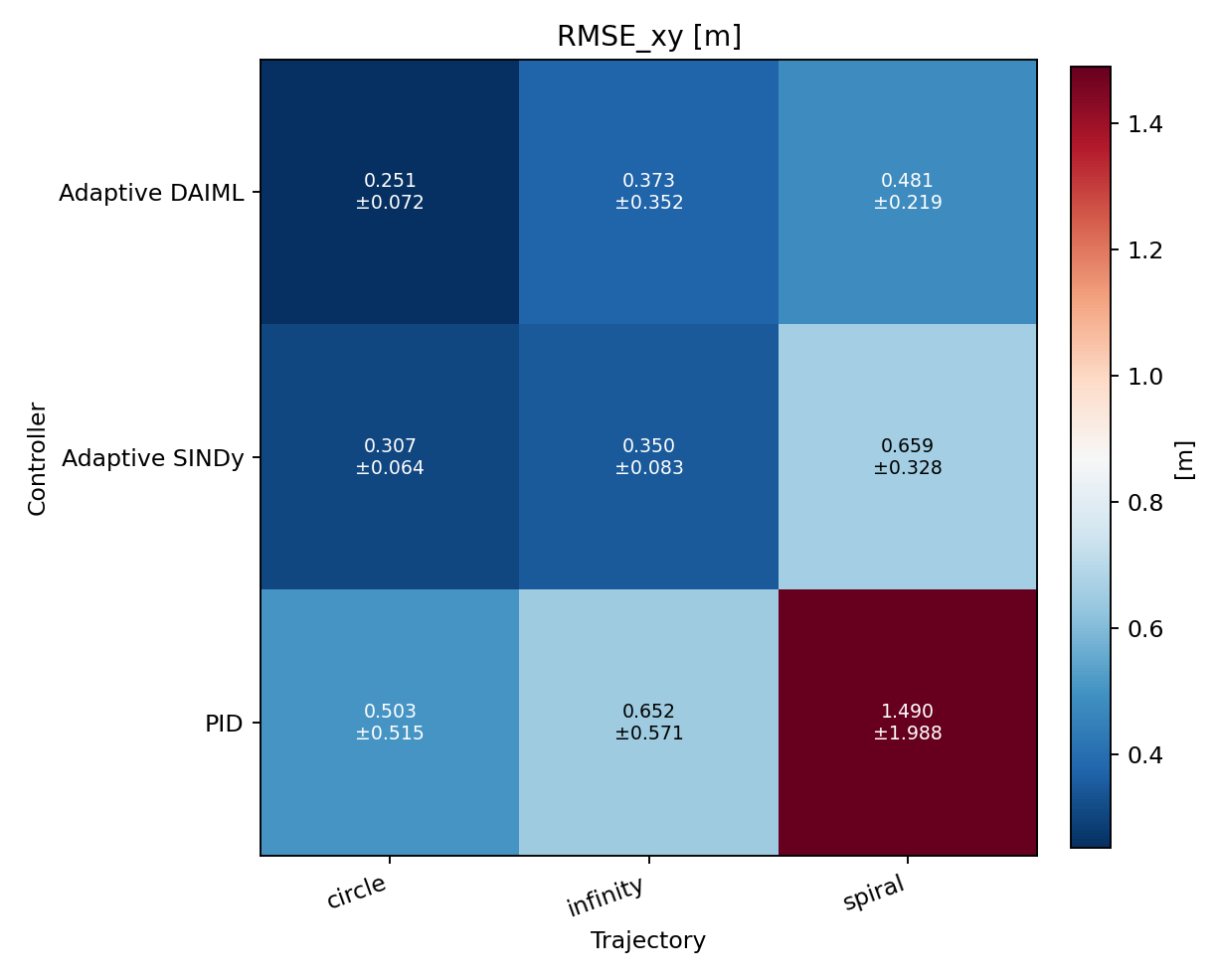}
    \caption{RMSE XY}
    \label{fig:a}
  \end{subfigure}
  \hfill
  \begin{subfigure}[b]{0.42\textwidth}
    \centering
    \includegraphics[width=\linewidth]{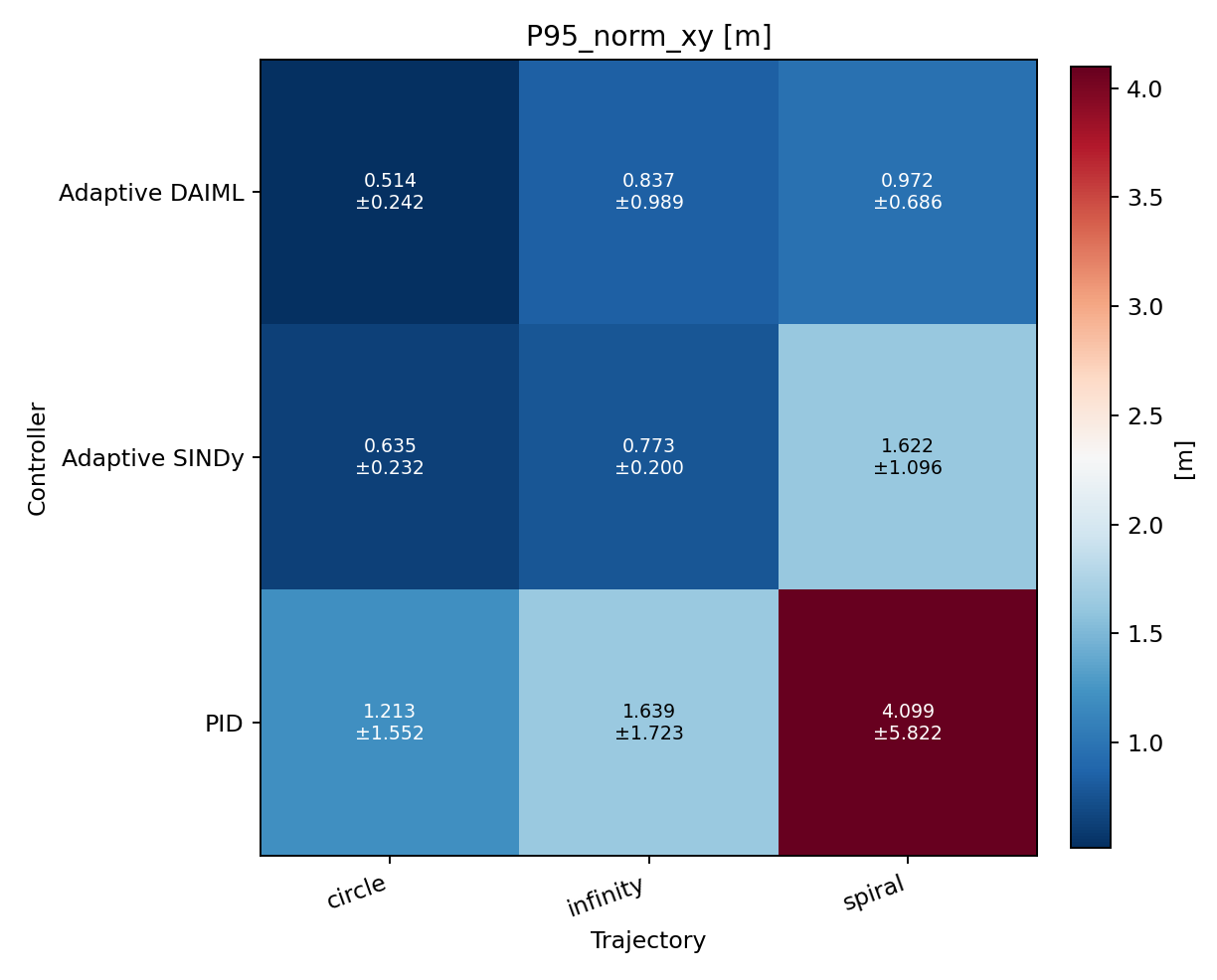}
    \caption{P95 Norm XY}
    \label{fig:d}
  \end{subfigure}
  
  \caption{Error Heatmap Comparison for ArduPilot non-convex trajectory tracking under wind disturbance in Gazebo simulation  }
  \label{fig:ardupilot_heatmap}
  \vspace{-0.8cm}
\end{figure}

All three trajectories with each controller were analyzed in 10 Gazebo simulation runs. Fig.~\ref{fig:ardupilot_heatmap} illustrates these comparisons over the three trajectories with a heatmap.   

For the circular trajectory, the adaptive approach clearly outperforms the baseline PID with reduced RMSE of 0.20-0.25 m. More importantly, the P95 norm shows a reduced value showing the concentration of error distribution under 0.50 m compared to 1.30 m for the baseline, which illustrates that even under worst scenarios the adaptive approach keeps the error under a low bound. The DAIML performs slightly better than Adaptive SINDy on the circular and spiral trajectories, but the performance is closely comparable. Similarly, for the lemniscate trajectory both the adaptive controllers perform much better than the baseline with Adaptive SINDy showing a slight improvement over the DAIML. The important thing to note from the heatmaps in Fig. \ref{fig:ardupilot_heatmap} and Table \ref{tab:ardupilot_error} is the distribution of the error statistics. For both adaptive approaches, the system keeps the errors within low bounds, the max errors stay within close vicinity of the RMSE, while PID often tends to go in a runaway situation where the drone is incapable of rejecting a sudden gust.
A more comprehensive summary of the results is shown in Table \ref{tab:ardupilot_error}.

% \begin{table*}[h]
% \centering
% \caption{Tracking error statistics across trajectories (mean $\pm$ std over 10 runs).}
% \label{tab:ardupilot_error}
% \scriptsize
% \setlength{\tabcolsep}{18pt}
% \begin{tabular}{|c|c|c|c|c|c|}
% \hline
% \textbf{Traj.} & \textbf{Ctrl.} & \textbf{RMSE$_{xy}$} & \textbf{MAE$_{xy}$} & \textbf{P95$_{\|xy\|}$} & \textbf{Max$_{\|xy\|}$} \\
%  &  & \textbf{[m]} & \textbf{[m]} & \textbf{[m]} & \textbf{[m]} \\
% \hline
% circle   & ADAIML & \textbf{0.251 $\pm$ 0.072} & \textbf{0.195 $\pm$ 0.029} & \textbf{0.514 $\pm$ 0.242 }& \textbf{0.851 $\pm$ 0.420} \\
% \hline
% circle   & \textbf{ASINDy} & 0.307 $\pm$ 0.064 & 0.234 $\pm$ 0.025 & 0.635 $\pm$ 0.232 & 1.170 $\pm$ 0.360 \\
% \hline
% circle   & PID    & 0.503 $\pm$ 0.515 & 0.343 $\pm$ 0.257 & 1.213 $\pm$ 1.552 & 1.658 $\pm$ 1.887 \\
% \hline
% infinity & ADAIML & 0.373 $\pm$ 0.352 & 0.267 $\pm$ 0.181 & 0.837 $\pm$ 0.989 & 1.395 $\pm$ 1.374 \\
% \hline
% infinity & \textbf{ASINDy} & \textbf{0.350 $\pm$ 0.083 }& \textbf{0.262 $\pm$ 0.054} & \textbf{0.773 $\pm$ 0.200} & \textbf{1.231 $\pm$ 0.340} \\
% \hline
% infinity & PID    & 0.652 $\pm$ 0.571 & 0.434 $\pm$ 0.296 & 1.639 $\pm$ 1.723 & 2.308 $\pm$ 2.059 \\
% \hline
% spiral   & ADAIML & \textbf{0.481 $\pm$ 0.219 }& \textbf{0.287 $\pm$ 0.108} & \textbf{0.972 $\pm$ 0.686} & \textbf{2.193 $\pm$ 0.804} \\
% \hline
% spiral   & \textbf{ASINDy} & 0.659 $\pm$ 0.328 & 0.373 $\pm$ 0.126 & 1.622 $\pm$ 1.096 & 2.774 $\pm$ 1.409 \\
% \hline
% spiral   & PID    & 1.490 $\pm$ 1.988 & 0.730 $\pm$ 0.821 & 4.099 $\pm$ 5.822 & 6.335 $\pm$ 9.368 \\
% \hline
% \end{tabular}
% \end{table*}

Across the ArduPilot SITL trials, adaptive SINDy shows its clearest advantage on the more challenging infinity (lemniscate) trajectory, where it slightly outperforms adaptive DAIML on all reported metrics. The achieved RMSE of 0.350 m vs 0.373 m, MAE 0.262 m vs 0.267 m, and P95 of 0.773 m vs 0.837 m indicate both improved typical tracking and a tighter error tail. While DAIML remains best on the circle and spiral cases, adaptive SINDy still consistently improves over the baseline PID on all trajectories, with substantially lower RMSE and percentile errors and reduced worst-case deviations. On a broader outlook, both the adaptive approaches perform very similar and have a clear advantage over non-adaptive PID.

\subsection{Crazyflie SITL Disturbance Rejection}

\begin{figure}[h!]
  \centering

  \begin{subfigure}[b]{0.42\textwidth}
    \centering
    \includegraphics[width=\linewidth]{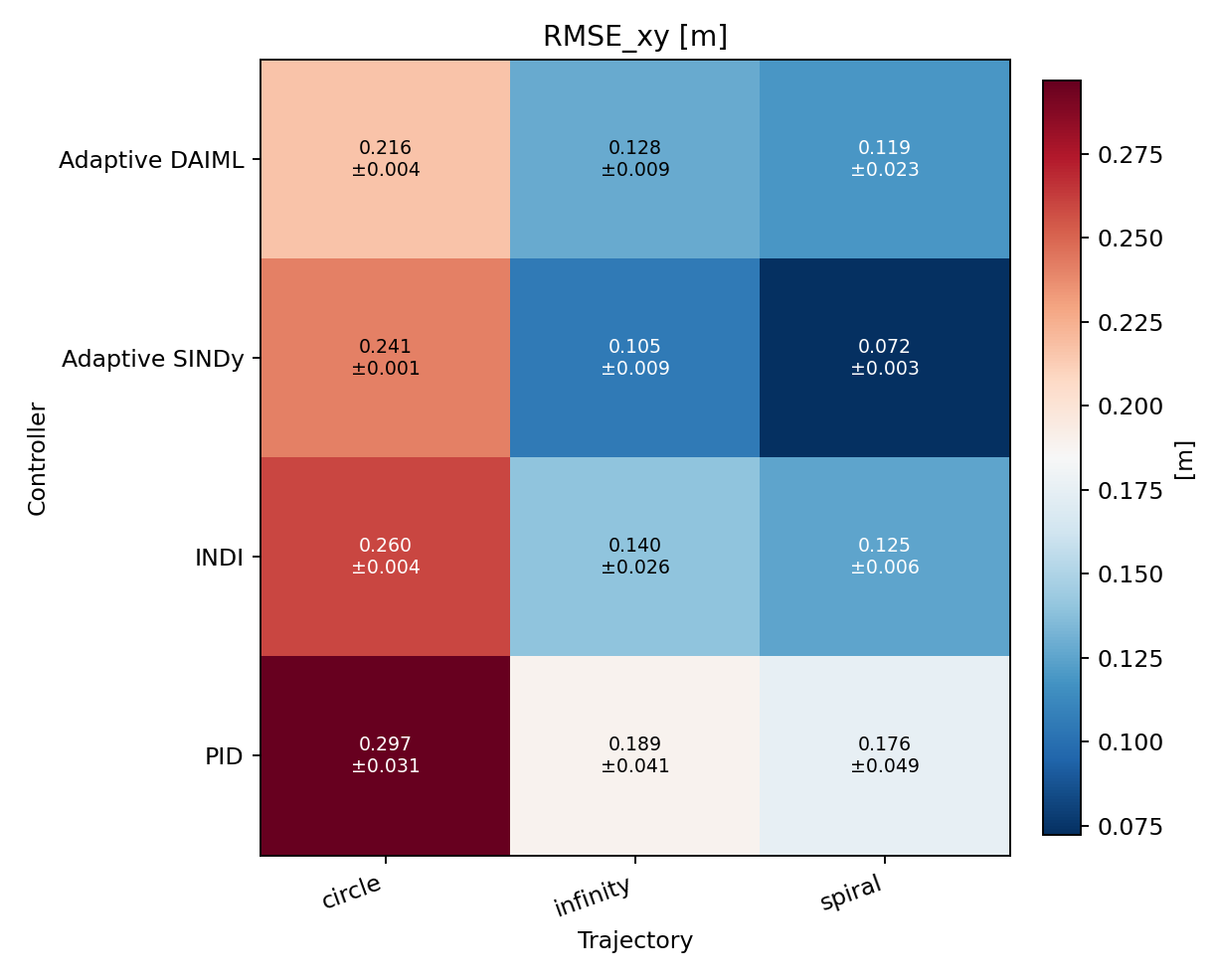}
    \caption{RMSE XY}
    \label{fig:a}
  \end{subfigure}
  \hfill
  \begin{subfigure}[b]{0.42\textwidth}
    \centering
    \includegraphics[width=\linewidth]{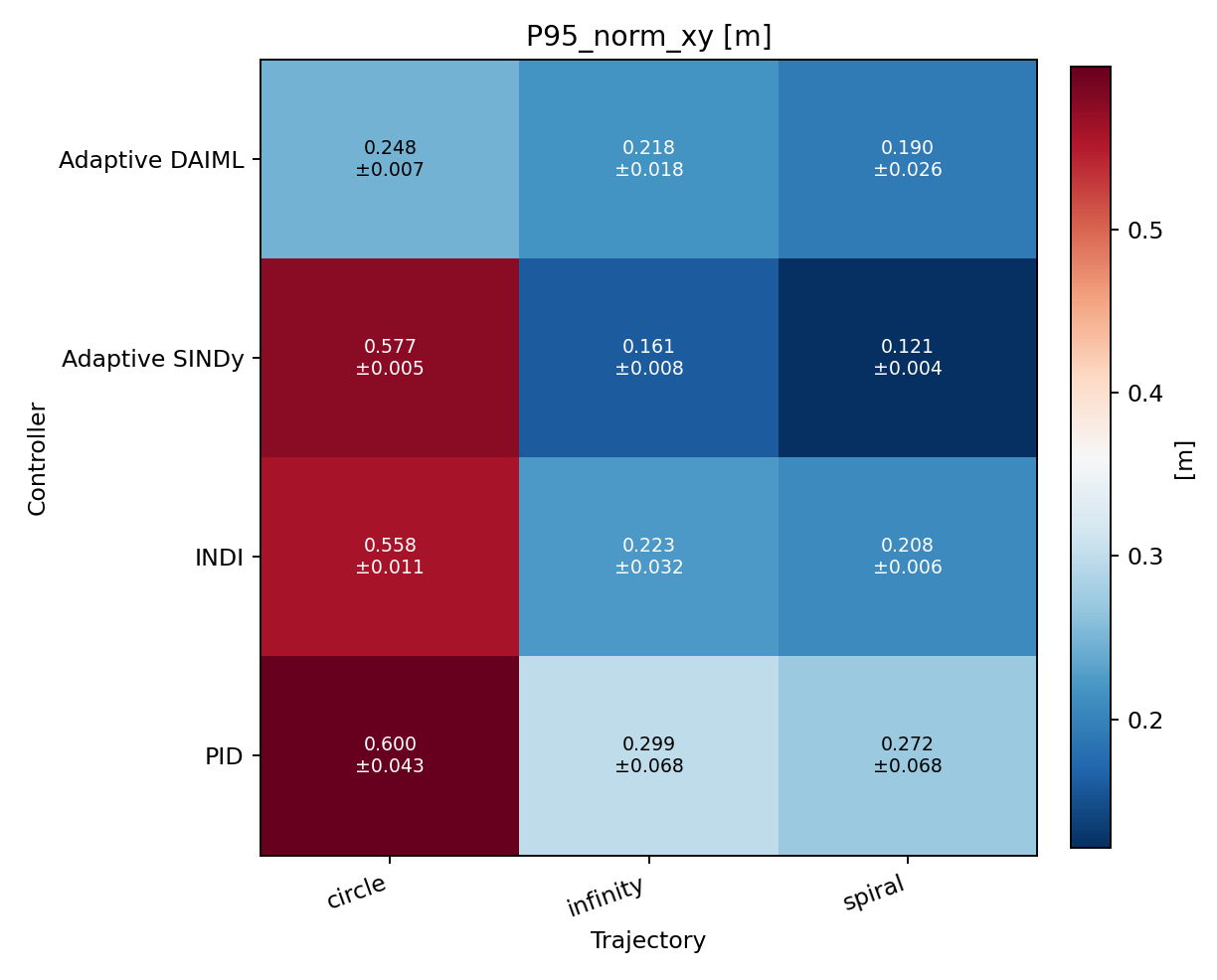}
    \caption{P95 Norm XY}
    \label{fig:d}
  \end{subfigure}

  \caption{Error Heatmap Comparison for Crazyflie non-convex trajectory tracking under disturbances in Gazebo simulation.}
  \label{fig:crazyflie_heatmap}
  \vspace{-0.5cm}
\end{figure}

The same simulation analysis was performed for a Crazyflie in Gazebo. This also served as a validation of the approaches over a different system with the same pre-trained DAIML and SINDy system identification-based adaptation subject to changes in tuning parameters. In addition to the three approaches considered in the previous section, we leveraged the ability to analyze an INDI controller on Crazyflie and added it to our analysis. Fig.~\ref{fig:crazyflie_heatmap} illustrates these comparisons over the three trajectories with a heatmap.  

On the Crazyflie system, Adaptive SINDy shows its strongest gains on the infinity and spiral trajectories. For the infinity trajectory, it achieves the lowest overall tracking errors among all controllers, reducing RMSE to 0.105 m, and tightening the tail to P95 0.161 m, outperforming DAIML, INDI, and PID. While on the circular trajectory, DAIML remains better in RMSE/P95, Adaptive SINDy still improves over PID and delivers competitive average errors, indicating that its advantage becomes more pronounced as the trajectory demands and disturbance-response complexity increase. Table \ref{tab:crazyflie_tracking_xy} summarizes the result for all the controllers and the trajectories. As compared to the ArduPilot setup, the overall errors are generally low, this is purely due to the size of the drone and the trajectory that it is following.

% \newpage
\subsection{Real Flight Results}

\begin{figure*}[t]
  \centering
  \begin{subfigure}[b]{0.32\textwidth}
    \centering
    \includegraphics[width=\linewidth]{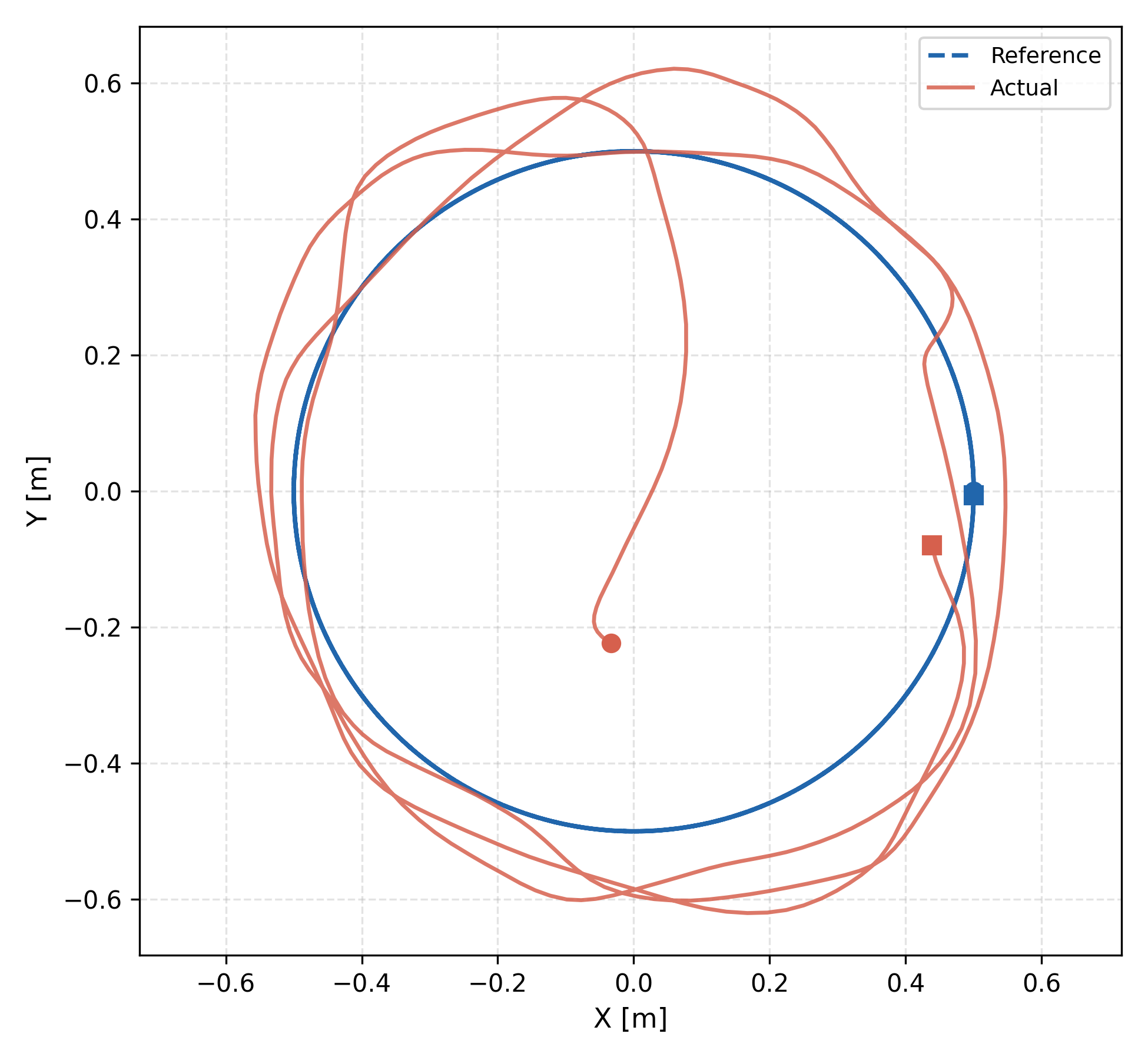}
    \caption{Circle}
    \label{fig:real_circle}
  \end{subfigure}
  \hfill
  \begin{subfigure}[b]{0.32\textwidth}
    \centering
    \includegraphics[width=\linewidth]{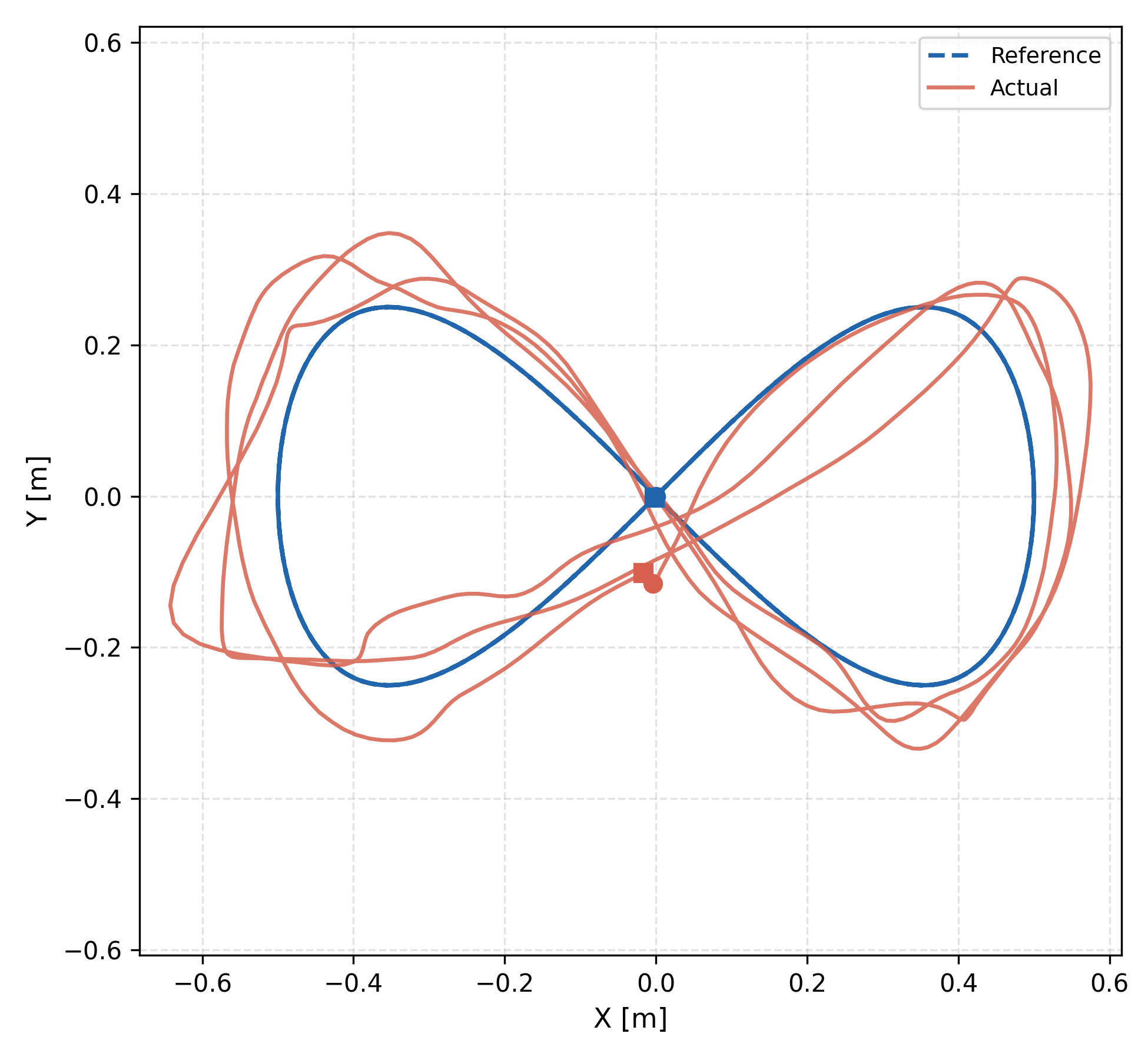}
    \caption{Lemniscate}
    \label{fig:real_lemniscate}
  \end{subfigure}
  \hfill
  \begin{subfigure}[b]{0.32\textwidth}
    \centering
    \includegraphics[width=\linewidth]{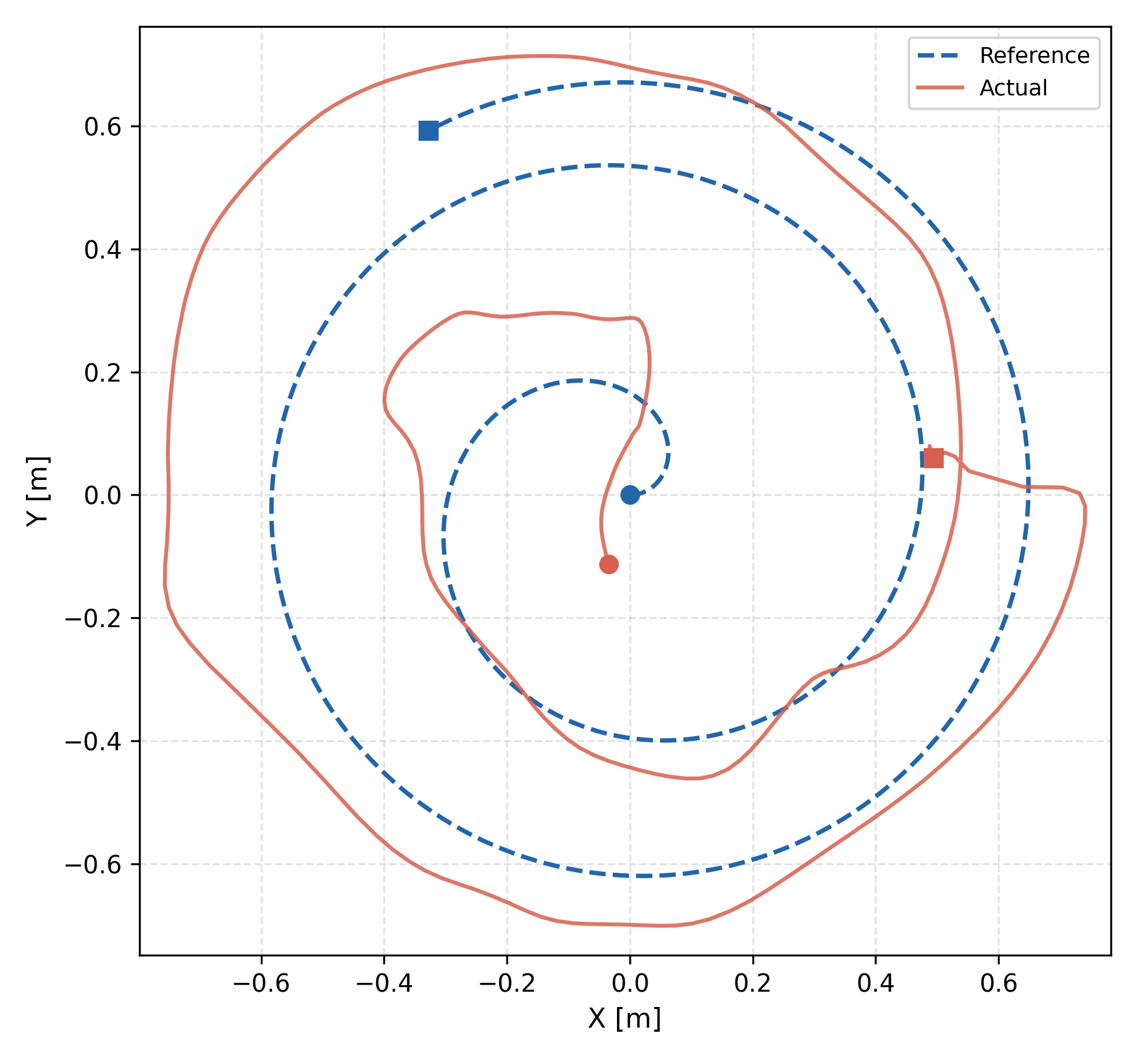}
    \caption{Spiral}
    \label{fig:real_spiral}
  \end{subfigure}

  \caption{Real flight trajectory overlays (blue dashed: reference, orange solid: achieved). The Adaptive SINDy controller tracks all three trajectory types. The oscillatory nature of the achieved trajectory illustrate the turbulent nature of the environment.}
  \label{fig:real_flight_traj}
\end{figure*}

\begin{table*}[h!]
\centering
\caption{Real-flight tracking error statistics for the Adaptive SINDy controller (mean $\pm$ std). The PID baseline crashed on all attempted runs and is excluded.}
\label{tab:real_flight}
\scriptsize
\setlength{\tabcolsep}{18pt}
\begin{tabular}{|c|c|c|c|c|c|}
\hline
\textbf{Trajectory} & \textbf{Runs} & \textbf{RMSE$_{xy}$} & \textbf{MAE$_{xy}$} & \textbf{P95$_{\|xy\|}$} & \textbf{Max$_{\|xy\|}$} \\
 &  & \textbf{[m]} & \textbf{[m]} & \textbf{[m]} & \textbf{[m]} \\
\hline
circle   & 6 & 0.176 $\pm$ 0.097 & 0.137 $\pm$ 0.060 & 0.379 $\pm$ 0.275 & 0.435 $\pm$ 0.305 \\
\hline
infinity & 4 & 0.122 $\pm$ 0.047 & 0.105 $\pm$ 0.038 & 0.234 $\pm$ 0.122 & 0.375 $\pm$ 0.166 \\
\hline
spiral   & 5 & 0.327 $\pm$ 0.199 & 0.255 $\pm$ 0.178 & 0.751 $\pm$ 0.394 & 0.873 $\pm$ 0.433 \\
\hline
\end{tabular}
\end{table*}

Real flight experiments were conducted on a Crazyflie 2.1 platform equipped
with a STM32F405 microcontroller.
Wind disturbances were generated using four ducted fans placed 2 meters away from the center of the flight zone as shown in Fig. \ref{fig:title}.  
The same three reference trajectories used in simulation i.e., circle, lemniscate, and spiral, were commanded to the vehicle, and states and reference
signals were logged at approximately 28~Hz.
The proposed Adaptive SINDy controller was evaluated over a total of 15 real flight
runs (6 circle, 4 lemniscate, 5 spiral).
A baseline PID controller was also tested under identical wind conditions.

The PID baseline failed to complete any trajectory in the presence of real wind
disturbances, resulting in an unrecoverable crash on every attempted run.
% (2 circle, 1 lemniscate, 1 spiral).
Consequently, no quantitative comparison with the PID baseline is possible; however,
the qualitative outcome itself constitutes a compelling validation of the necessity
of system identification based adaptive disturbance rejection.
The Adaptive SINDy controller successfully completed all 15 runs across all three
trajectory types with no failures.
Fig.~\ref{fig:real_flight_traj} shows representative trajectory overlays for each
trajectory type.
A full per-trajectory breakdown is given in Table~\ref{tab:real_flight}.

% \begin{table}[h]
% \centering
% \caption{Real-flight tracking error statistics for the Adaptive SINDy controller (mean $\pm$ std). The PID baseline crashed on all attempted runs and is excluded.}
% \label{tab:real_flight}
% \scriptsize
% \setlength{\tabcolsep}{3pt}
% \begin{tabular}{|c|c|c|c|c|c|}
% \hline
% \textbf{Traj.} & \textbf{Runs} & \textbf{RMSE$_{xy}$} & \textbf{MAE$_{xy}$} & \textbf{P95$_{\|xy\|}$} & \textbf{Max$_{\|xy\|}$} \\
%  &  & \textbf{[m]} & \textbf{[m]} & \textbf{[m]} & \textbf{[m]} \\
% \hline
% circle   & 6 & 0.176 $\pm$ 0.097 & 0.137 $\pm$ 0.060 & 0.379 $\pm$ 0.275 & 0.435 $\pm$ 0.305 \\
% \hline
% infinity & 4 & 0.122 $\pm$ 0.047 & 0.105 $\pm$ 0.038 & 0.234 $\pm$ 0.122 & 0.375 $\pm$ 0.166 \\
% \hline
% spiral   & 5 & 0.327 $\pm$ 0.199 & 0.255 $\pm$ 0.178 & 0.751 $\pm$ 0.394 & 0.873 $\pm$ 0.433 \\
% \hline
% \end{tabular}
% \end{table}

The lemniscate trajectory yields the lowest errors
(RMSE$_{xy}$ 0.122 $\pm$ 0.047~m, P95 0.234 $\pm$ 0.122~m), consistent with
its smooth and symmetric curvature.
The circle trajectory shows moderate errors
(RMSE$_{xy}$ 0.176 $\pm$ 0.097~m), and the spiral trajectory presents the
greatest challenge (RMSE$_{xy}$ 0.327 $\pm$ 0.199~m, P95 0.751 $\pm$ 0.394~m)
due to its continuously expanding radius and the correspondingly increasing demand on the disturbance estimator. The experimental validation demonstrates both the strength of our approach and the sim to real transfer. The turbulent nature of the environment can be inferred from the graphs of Fig. \ref{fig:real_flight_traj} that show successful tracking of all three trajectories without crashing.

\newpage
\section{Conclusion and Future Work}

In this study, we implemented a novel approach of SINDy-based system identification of wind induced residual forces on a quadcopter UAV. The identification of system dynamics was coupled with an RLS adaptive controller to estimate residual forces and actively reject the disturbance in a turbulent environment. A physics-informed library for basis functions was used using the observation of attitude changes and thrust variation on the drone under the influence of the wind. The identified representation of the dynamics was used in the adaptive control to simultaneously update the adaptation parameters. The architecture was implemented in gazebo on ArduPilot SITL and Crazyflie platforms and experimentally validated on real Crazyflie flights. Adaptive SINDy outperformed baseline PID control on all non-convex trajectories, achieving a low RMSE of 17.6 cm, MAE of 13.7 cm on circular trajectory, and RMSE of 12.2 cm, MAE of 10.5 cm on lemniscate trajectory. Moreover, the validation strength is depicted by the implementation of the architecture on a lightweight system susceptible to runaway conditions; our method successfully completed all real flights under a turbulent environment with winds applied from four directions.

This work lays a foundation for the hybrid of classical adaptive control with data-driven system identification for UAVs. 
% Although the experimental validation is promising,
However, the potential of system identification under more turbulent environment for different types and sizes of UAV systems can be further explored. Additionally, system identification using relatively low amounts of data from state history can be implemented on-board to identify changing dynamics and adapt accordingly.  

\newpage
% \section*{Acknowledgements} 
% Research reported in this publication was financially supported by the RSF grant No. 24-41-02039.

% \nocite{*}

\bibliographystyle{IEEEtran}
\bibliography{bibliography}
\balance
\addtolength{\textheight}{-12cm}
\end{document}